\documentclass[10pt,twocolumn,letterpaper]{article}

\usepackage{cvpr}              % To produce the CAMERA-READY version
\definecolor{cvprblue}{rgb}{0.21,0.49,0.74}
\usepackage[pagebackref,breaklinks,colorlinks,allcolors=cvprblue]{hyperref}
\usepackage{multirow}
\usepackage{float}

\def\paperID{***} % *** Enter the Paper ID here
\def\confName{CVPR}
\def\confYear{2026}

\title{HumanOrbit: 3D Human Reconstruction as 360° Orbit Generation}

\author{Keito Suzuki\textsuperscript{*}\textsuperscript{$\dagger$} \quad Kunyao Chen\textsuperscript{$\dagger$} \quad Lei Wang\textsuperscript{$\dagger$} \quad Bang Du\textsuperscript{*} \quad Runfa Blark Li\textsuperscript{*} \\
Peng Liu\textsuperscript{$\dagger$} \quad Ning Bi\textsuperscript{$\dagger$} \quad Truong Nguyen\textsuperscript{*} \\
\textsuperscript{*}University of California, San Diego \quad \textsuperscript{$\dagger$}Qualcomm
}

\begin{document}
\twocolumn[{%
\renewcommand\twocolumn[1][]{#1}%
\maketitle
\begin{figure}[H]
\hsize=\textwidth
\centering
\includegraphics[width=0.98\textwidth]{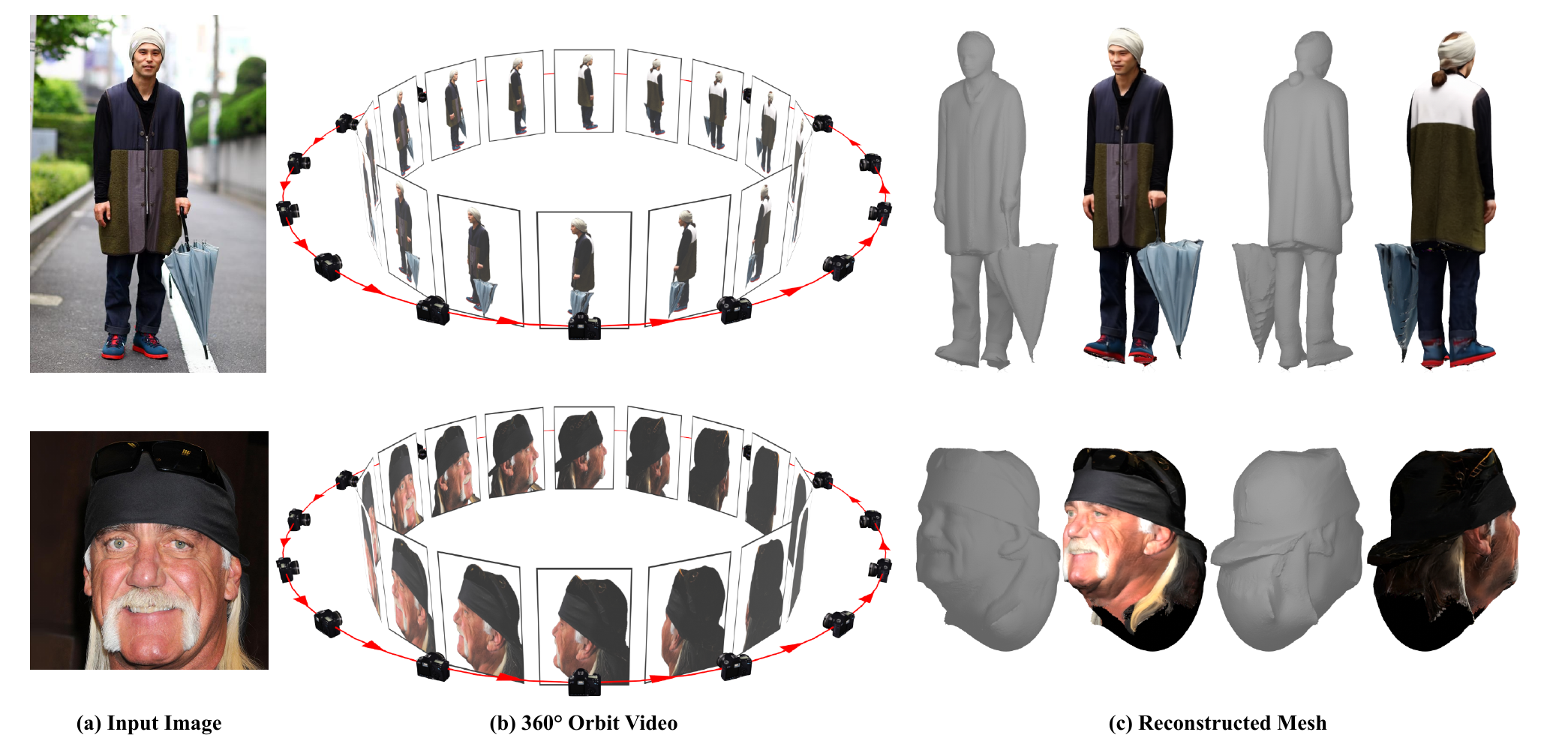}
\vspace{-2ex}
% \caption{Given a single image, our model directly generates a 360° orbit video around the subject. Using the generated frames, we propose a pipeline to reconstruct a textured mesh. Our results show that our method is versatile and can be applied to various human related images.
% }
\caption{Given an in-the-wild image, we treat the ill-posed problem of multi-view synthesis as orbit video generation for robust 3D human reconstruction. Our model creates identity preserving and view consistent frames, resulting in a high quality textured mesh. 
}
\label{Fig.cover}
\end{figure}
}]

\begin{abstract}
% We present a method for generating a full 360° multi-view images of a person from a single input image by finetuning a video diffusion model. Our approach enables the diffusion model to synthesize continuous camera rotations around the subject, producing geometrically consistent novel views while preserving the appearance and identity of the person. Using the generated multi-view frames, we further propose a reconstruction pipeline that recovers a high-quality 3D mesh of the subject. Experimental results demonstrate that our camera-orbit LoRA effectively extends a pre-trained video diffusion model for multi-view image generation and that the reconstructed 3D models exhibit superior completeness and fidelity compared to those from state-of-the-art baselines.
We present a method for generating a full 360° orbit video around a person from a single input image. Existing methods typically adapt image-based diffusion models for multi-view synthesis, but yield inconsistent results across views and with the original identity. In contrast, recent video diffusion models have demonstrated their ability in generating photorealistic results that align well with the given prompts. Inspired by these results, we propose HumanOrbit, a video diffusion model for multi-view human image generation. Our approach enables the model to synthesize continuous camera rotations around the subject, producing geometrically consistent novel views while preserving the appearance and identity of the person. Using the generated multi-view frames, we further propose a reconstruction pipeline that recovers a textured mesh of the subject. Experimental results validate the effectiveness of HumanOrbit for multi-view image generation and that the reconstructed 3D models exhibit superior completeness and fidelity compared to those from state-of-the-art baselines.
\end{abstract}    
\vspace{-3ex}
\section{Introduction}
\label{sec:intro}

Reconstructing a photorealistic 3D avatar from a single image is a long-standing problem with intriguing applications in areas such as telecommunication, gaming, and AR/VR. The task is fundamentally under-constrained, requiring recovery of 3D shape and appearance from one view despite variation in pose and clothing and extensive self-occlusion.

Recently, large reconstruction models \cite{lrm} have shown impressive capabilities in generating a 3D asset from a single image by leveraging large-scale datasets \cite{objaverse, mvimgnet}. However, they still lack the ability in reconstructing a photorealistic 3D human due to the significant lack of 3D human data. Human-specific models \cite{weng2024template} trained on publicly available 3D human data have also been proposed, yet their generalization ability still remains limited. 

In order to train a generalizable feed-forward model for 3D human reconstruction, a high quality multi-view or 3D dataset of diverse humans is essential. However, collecting such datasets remains costly and logistically demanding, often requiring specialized capture studios equipped with dense calibrated camera arrays, lighting, and controlled environments. As a result, this data collection scheme is difficult to scale with existing datasets covering only a limited diversity of subjects, clothing styles, and poses. 

In contrast to 3D data, 2D human images are abundant with various publicly available sources  \cite{CCP, deepfashion1, CelebAMask-HQ}. These large scale image collections capture the diversity that existing 3D datasets lack, presenting a promising direction for scalable 3D human data collection—if they can be effectively leveraged. While recent methods \cite{pshuman, meat} have adopted image-based diffusion models for multi-view synthesis, they still fail in generating coherent details and identities across views.  

In terms of consistency, recent video diffusion models \cite{stablediffusion, wan, hunyuanvideo} have shown remarkable capabilities in producing temporally coherent and photorealistic human videos, capturing complex motions, camera trajectories, and viewpoint changes with high realism. These models learn strong priors of 3D structure and appearance consistency across frames, suggesting their potential as powerful tools for generating novel views of a person.
Recently, Human4DiT \cite{human4dit} introduced 4D diffusion transformers for 4D human content generation, but their model still requires a large multi-dimensional dataset.

Inspired by these advances, we propose HumanOrbit, a video diffusion model that generates a 360° camera orbit video around the target subject from an input image. 
Our approach requires training only a small set of learnable parameters and achieves this capability using a modest training set of renderings from only 500 3D human scans, making it highly data-efficient. Importantly, our method is pose-free—it does not rely on external body pose or camera pose annotations, yet learns to produce smooth, 3D-consistent orbiting camera motion directly on single input images. 
Our approach is simple but still robust in generating dense and coherent multi-view images of diverse human cases due to the learned prior from billions of real-world video data. Experimental results show that our model outperforms recent multi-view diffusion models in generating 3D consistent and identity preserving multi-view images.

Using the generated multi-view images, we further propose a 3D reconstruction pipeline to obtain a textured mesh. Specifically, we apply a SfM method and normal estimation model to estimate the camera parameters and normal maps respectively for each frame. These results are then utilized to optimize a mesh via a fast mesh carving method \cite{continuous}. Through this process, we demonstrate that our orbit videos can be used to generate high quality 3D meshes.

In Figure \ref{Fig.cover}, we show some results from our model exhibiting high versatility in generating dense multi-view images for 3D reconstruction of the full body and the head. 

% Additional experiments with the same model on non-human images such as animals, show that our model can still generate view-consistent results, exhibiting a promising direction beyond human data generation.

To summarize, our main contributions are as follows:
\begin{itemize}
    \item We introduce HumanOrbit, a data-efficient video diffusion model that generates a high-fidelity 360° orbit video of a static subject from an input image. 
    \item We propose a 3D reconstruction pipeline to obtain a textured mesh from the generated pose-free video.
    \item Experimental results show that our method achieves state-of-the-art results in generating view-consistent frames and shows robustness in handling various human images.
\end{itemize}
\section{Related Works}
\subsection{Multi-view Diffusion Model}
\noindent
\textbf{General Image Synthesis.}
Recent trends in multi-view diffusion models were kick started with Zero-1-to-3 \cite{zero123} which proposed a viewpoint conditioned diffusion model for novel view synthesis. The main challenge in generating novel views has been maintaining the geometric and color consistency in the synthesized images. 
SyncDreamer \cite{syncdreamer} proposes a 3D-aware attention mechanism to synchronize the states in each view for better consistency. MVDream \cite{mvdream} introduces camera embeddings and an inflated 3D self-attention. EpiDiff \cite{epidiff} leverages epipolar constraints for cross-view interactions between features maps from nearby views. Era3D \cite{era3d} also uses epipolar priors with an efficient row-wise attention. Despite promising results on general objects, these models still exhibit artifacts when generalized to diverse human appearances.

\begin{figure*}[t]
    \centering
    \includegraphics[width=.98\hsize]{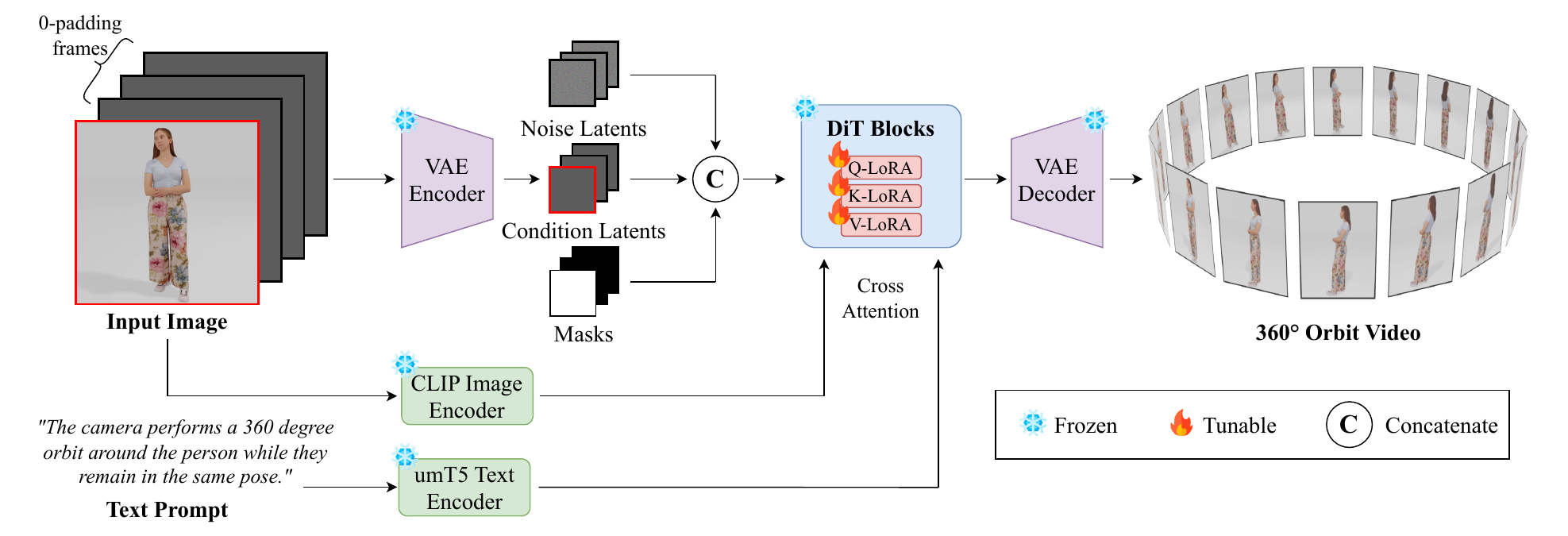}
    \vspace{-2ex}
    \caption{The proposed HumanOrbit model. We finetune a DiT-based video diffusion model such that it directly generates a 360° orbit video given a single input image. While keeping much of the architecture frozen, the finetuned LoRAs learn to accurately conduct an orbit around the subject, generating consistent multi-view images.} 
    \label{Fig.wan_model}
    \vspace{-2ex}
\end{figure*}

\noindent
\textbf{Human Image Synthesis.} 
MEAT \cite{meat} enables high-resolution multi-view generation by introducing a novel mesh attention that directly establishes cross-view correspondences. Pippo \cite{pippo} is a multi-view diffusion transformer \cite{dit} that generates turnaround videos conditioned on target camera views and spatial anchors. However, it requires multi-stage training and post-training on studio capture data. Human4DiT \cite{human4dit} can generate multi-view human videos, but it also requires collecting a large human dataset.

Our approach in multi-view human image generation is motivated by the impressive results in recent video diffusion models \cite{wan, cogvideox, svd, hunyuanvideo}. They demonstrate the ability to generate temporally coherent videos with diverse camera motions which hint at their capability in achieving view consistency in 3D space. Building on this insight, we repurpose a pretrained video diffusion model as a multi-view generator by finetuning on minimal 3D human data.

\subsection{Single Image 3D Human Reconstruction}
Reconstructing a 3D clothed human has gained significant popularity since the seminal work, PIFu \cite{pifu}, which proposed a data-driven method with pixel aligned features. PIFuHD \cite{pifuhd} extended this work to recover detailed geometry with higher resolution and normal guidance. Works such as ICON \cite{icon} and PaMIR \cite{pamir} incorporated parametric body priors \cite{smpl, smplx} to better regularize implicit functions for handling complex poses and clothing. PSHuman \cite{pshuman} also uses body priors but as an initialization for explicit mesh carving \cite{continuous}. Similar to PSHuman, our method also iteratively optimizes a mesh but does not require any pose and can be generalized to various human images.
\section{Proposed Method}
In this section, we first provide details about the proposed HumanOrbit model for dense multi-view image generation and then further explain the proposed pipeline to reconstruct a 3D mesh from the generated images.

\subsection{Omni-View Image Generation}
Recent works \cite{pshuman, meat} have mainly adapted image-based diffusion models \cite{stablediffusion, zero123} with 3D human body priors \cite{smpl, smplx} for multi-view human image generation. While they introduce novel attention mechanisms to tackle the 3D consistency problem, the inconsistency is still evident. These methods are still unable to achieve consistent details in areas such as the face and hands, and requires postprocessing for detailed 3D reconstruction. 
Furthermore, the reliance on parametric models restricts the use case to scenarios in which the full body is visible in the original image.

In contrast, recent DiT-based \cite{dit} video diffusion models have shown remarkable performance in generating realistic videos from texts and images. These models are trained on billions of real-world videos to generate complex animations with high temporal consistency. Inspired by their ability in generating coherent videos with various camera motions, we base our model on a pre-trained video diffusion model such that it consistently generates an orbit motion around a static person. Compared to previous methods, our model captures much denser multi-view images.

\noindent
\textbf{Model Architecture}
The structure of our model is shown in Figure \ref{Fig.wan_model}, comprised of a 3D VAE, an image encoder, a text encoder, and DiT blocks. Given an image from a single view, our model automatically generates dense multi-view images $\mathcal{I}=\{I_i\}_{i=1}^K$ where $K$ is the number of output frames.
Our model does not require additional inputs such as body pose or camera parameters.

The input image is first zero-padded temporally and passed through a VAE Encoder to get the condition latents. These latents are then concatenated with the noise and binary masks, which indicate the reference input frame, to be denoising by a series of DiT blocks. In each block, the latents interact with image and text embeddings, extracted from the input image and text prompt respectively, via cross attention. The denoised latents are finally decoded by the VAE Decoder to obtain a 360° orbit video.

\begin{figure*}[t]
    \centering
    \includegraphics[width=.98\hsize]{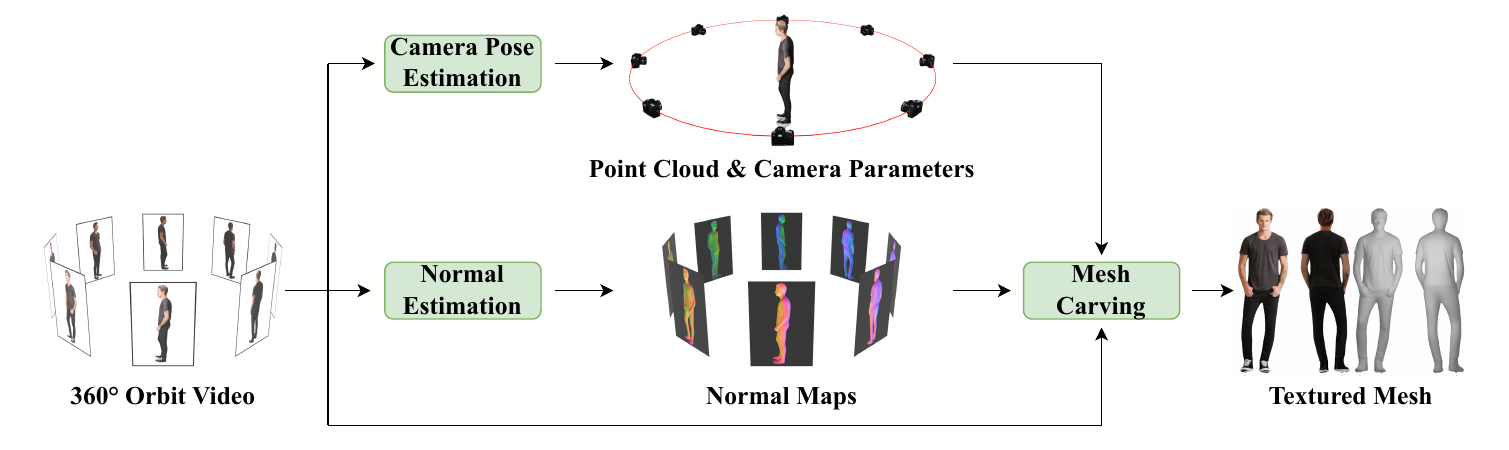}
    \vspace{-2ex}
    \caption{The proposed mesh reconstruction framework. Given the generated multi-view images, we first apply a SfM method to obtain the point cloud and camera parameters for each view.
     We then estimate the normal maps for each frame. Finally, using the previously generated results, the textured mesh is reconstructed via an explicit mesh carving method.} 
    \label{Fig.mesh_reconstruction}
\end{figure*}

\noindent
\textbf{Training.}
In order to effectively finetune a large video model for multi-view generation, we employ Low-Rank Adaptation (LoRA) \cite{lora} in the DiT blocks, as shown in Figure \ref{Fig.wan_model}. This allows us to train the model on a small video dataset without losing the generalization capability of the base video generation model.

To create the training dataset, we render orbital videos from 500 3D scans of the PosedPro dataset \cite{RenderPeople} using Blender. For each 3D scan, we render sequences that cover the full body as well as sequences that include only above the shoulders. We also add slight rotations as augmentation such that the subject is not always facing the camera in the initial frame. This results in a training dataset of 3,000 videos each containing $K$ frames at a resolution of $640\times640$. Each training sample is also paired with the same text prompt as shown in Figure \ref{Fig.wan_model}.
Examples of the training data is shown in the supplementary.

\subsection{3D Mesh Reconstruction}
Given the generated multi-view images, we propose a framework to reconstruct a textured mesh. This is illustrated in Figure \ref{Fig.mesh_reconstruction}. Similar to recent 3D mesh generation works \cite{pshuman, unique3d, craftsman3d, infinihuman} we use an optimization approach \cite{continuous} to reconstruct a mesh. This approach utilizes differentiable rendering and iteratively conducts vertex displacement and remeshing to get the optimized mesh. However, this requires accurate camera pose as well as normal maps for each multi-view image.

\begin{figure}[t]
    \centering
    \includegraphics[width=.98\hsize]{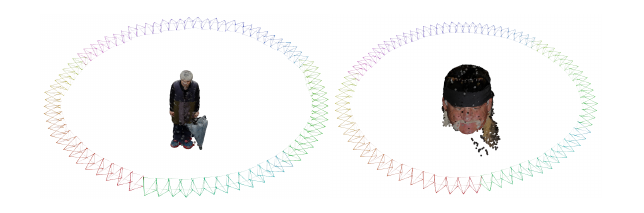}
    \vspace{-2ex}
    \caption{Two examples of the camera pose and point cloud predicted by VGGT on our generated multi-view images. The original images are shown in Figure \ref{Fig.cover}.} 
    \label{Fig.vggt_cameras}
\end{figure}

\noindent
\textbf{Camera Pose Estimation.} 
Previous methods rely on predefined camera poses or allow additional inputs to generate images from those corresponding views. However, there is no guarantee that the generated images align with the original camera poses. Therefore, to obtain accurate camera parameters for each frame, we apply an SfM method. Specifically, we use VGGT \cite{vggt}, a state-of-the-art feed-forward neural network for estimating 3D scene attributes from input images. We directly feed our generated multi-view images to VGGT to obtain the camera parameters $\Pi=\{\pi_i\}_{i=1}^K$, as well as the point cloud projected from the depth maps. 
Examples of these results are shown in Figure \ref{Fig.vggt_cameras} where it shows that VGGT is able to reliably predict the circular camera trajectory and initial point cloud from our generated images. 
It also further supports the view consistency in our multi-view images as view inconsistencies would lead to poor point cloud results. In the supplementary materials, we show that with previous multi-view generation methods, VGGT is unable to obtain a complete point cloud due to the low confidence in the estimated points.
% Although VGGT is able to output reasonable camera poses, the results are still unsatisfactory for 3D reconstruction. Therefore, we refine the camera pose through bundle adjustment. To compute the feature tracks, we uniformly sample query points within a segmentation mask in each query frame and then apply the VGGSfM tracker to track across views. This ensures complete coverage of the entire body as opposed to using keypoints which typically only covers edges and areas with texture. 

\noindent
\textbf{Mesh Carving.} After collecting the RGB frames $\mathcal{I}$,  we compute the normal maps $\mathcal{N}=\{N_i\}_{i=1}^K$ with an off-the-shelf normal estimation method. Using these results and the camera parameters $\Pi$ we reconstruct the mesh via differentiable rendering. We first initialize the mesh by applying Poisson Surface Reconstruction \cite{poisson} on the point cloud predicted by VGGT. Previous methods \cite{pshuman, infinihuman} have adopted fitted SMPL models \cite{smpl} as the initial mesh, but this limits the use case to full body human reconstruction. To maintain the generalizability beyond full body scenarios, we propose this robust initialization strategy. 

Given a mesh with vertices $V$ and faces $F$, we refine the details through a differentiable renderer $\mathcal{R}(V, F, \pi_i)$ which yields a mask $\hat{M}_i$ and normal map $\hat{N}_i$ for each view $i$.  During each optimization step, we optimize a combination of a mask loss:
\begin{equation}
    \mathcal{L}_{mask} = \sum_i \left\|M_i-\hat{M}_i \right\|_2^2
\end{equation}
where $M_i$ is the binary mask predicted from a segmentation model, and a normal loss:
\begin{equation}
    \mathcal{L}_{normal} = \sum_i M_i \odot \left\|N_i-\hat{N}_i \right\|_2^2
\end{equation}
where $\odot$ denotes the element-wise product. The final loss is computed as:
\begin{equation}
    \mathcal{L}_{recon} = \mathcal{L}_{mask} + \mathcal{L}_{normal}
\end{equation}

Once we got the refined geometry, we optimize the per-vertex colors by minimizing:
\begin{equation}
    \mathcal{L}_{color} = \sum_i M_i \odot \left\|I_i-\hat{I}_i \right\|_2
\end{equation}
where $\hat{I}_i$ is the rendered RGB image from view $i$.
\section{Experiments}
\subsection{Implementation Details}
For the base video diffusion model, we adopt the Wan 2.1 Image-to-Video 480p model \cite{wan}. This model uses the Wan-VAE, CLIP image encoder \cite{clip}, and umT5 text encoder \cite{umt5} for the VAE, image encoder, and text encoder respectively. Following this model, HumanOrbit generates $K=81$ frames.
We set the LoRA rank to 32 and train the model for 10 epochs on a single A100 GPU. 
For the normal estimation, we adopt NormalCrafter \cite{normalcrafter} as it can obtain temporally consistent normal maps from an input video.

\subsection{Multi-view Image Generation}

\noindent
\textbf{Datasets.}
To showcase our model's ability in generating consistent multi-view images for both full body and head portraits, we do a comparison on two datasets. For the full body analysis, we use 100 images from the Clothing Co-Parsing (CCP) dataset \cite{CCP}. For the head portraits, we also use 100 images from the CelebAMask-HQ dataset \cite{CelebAMask-HQ}.

\noindent
\textbf{Baselines.}
For both full body and head cases, we compare the performance with SV3D \cite{sv3d} and MV-Adapter \cite{mv-adapter}.
SV3D is a recent video diffusion model for generating orbital videos with 21 frames around a subject where we use the `SV3D\_u' version. MV-Adapter applies a plug-and-play adapter to text-to-image models for efficiently repurposing them for multi-view generation. We use their SDXL \cite{sdxl} model.
In the case of full body multi-view image generation, we also do a comparison with PSHuman \cite{pshuman} which uses cross-scale multi-view diffusion model to better preserve the identity the generated views. Since both MV-Adapter and PSHuman generates views from the same six angles, we select similar view angles for SV3D and our method when conducting numerical evaluation.
We did not conduct comparisons with some recent multi-view human image generation works \cite{spinmeround, meat, pippo, human4dit} since their code was not publicly available.
 
\noindent
\textbf{Evaluation Metrics.} Evaluating multi-view image generation from a single image is inherently ill-posed as there are multiple plausible results that can be generated from the input prompt. Consequently, simply measuring the photometric accuracy between the generated views and the ground truth images from a multi-view or 3D datasets is unreliable, as mentioned in previous works \cite{fancy123}.
We therefore adopt three different metrics: CLIP Score \cite{clip}, MEt3R \cite{met3r}, and MVReward \cite{mvreward}. These metrics operate without ground truth multi-view images and measure the coherence between the input prompt and the generated views as well as the 3D consistency between the synthesized views.

\noindent
\textit{CLIP Score} \cite{clip} is used to measure the discrepancy between the image prompt and the generated views. We use the pretrained CLIP image encoder $f_{I}$ to obtain the embeddings for the input image $I_{p}$ as well as each generated view $I_v$. The CLIP Score is computed by averaging the cosine similarity between the input image embedding and each generated view embedding:
\begin{equation}
    \text{CLIP Score} = \frac{1}{K}\sum_{i=1}^{K}  \frac{f_I(I_{p})\cdot f_I(I_{v})}{||f_I(I_{p})||||f_I(I_{v})||}
\end{equation}
where $K$ is the total number of generated views.

\noindent
\textit{MEt3R} \cite{met3r} quantifies the 3D consistency between the generated images. Given a pair of images, MEt3R obtains feature maps for each image using DINO \cite{dino, dinov2} and FeatUp \cite{featup}, while also computing a dense 3D reconstruction with DUSt3R \cite{dust3r}. It then projects the feature maps from one frame to the other and calculates the  similarity between the obtained feature maps. For our purpose, we compute the average MEt3R score between consecutive views.

\noindent
\textit{MVReward} \cite{mvreward} is a recently proposed model that encodes human preferences to evaluate multi-view diffusion models. The model takes generated RGB images and normal maps as input and directly outputs a reward score in which a higher value suggests a higher quality and consistency in the generated images. For methods that do not output normal maps, we compute them using NormalCrafter \cite{normalcrafter}.

\begin{table}
    \centering
    \caption{Quantitative comparison of multi-view image generation results on in-the-wild human images. For full body and head image evaluation we use the CCP \cite{CCP} and CelebA \cite{CelebAMask-HQ} dataset respectively. Results in \textbf{bold} show the best value.}
    \label{tab:numerical_comparison}
    \resizebox{0.98\columnwidth}{!}{\begin{tabular}{c|c|ccc}
    Dataset & Method & CLIP Score \cite{clip} $\uparrow$ & MEt3R \cite{met3r} $\downarrow$ & MVReward \cite{mvreward} $\uparrow$\\ \hline
    \multirow{4}{*}{Full body \cite{CCP}} & SV3D &  0.7888 & \textbf{0.2966} & 0.2378\\
    & MV-Adapter & 0.7735 & 0.3721 & 0.6795\\
     & PSHuman &  0.8282 & 0.3576 & 0.6814\\ 
     & Ours &  \textbf{0.8317} & 0.3175 & \textbf{0.8035}\\ \hline
    \multirow{3}{*}{Head \cite{CelebAMask-HQ}} & SV3D &  0.6582 & 0.4745 & 0.4918 \\ 
    & MV-Adapter & 0.6729 & 0.4826 & 0.4727\\
     & Ours &  \textbf{0.7073} & \textbf{0.4176} & \textbf{0.4947}\\
    \end{tabular}
    }
    \vspace{-3ex}
\end{table}

\begin{figure*}[t]
    \centering
    \includegraphics[width=.98\hsize]{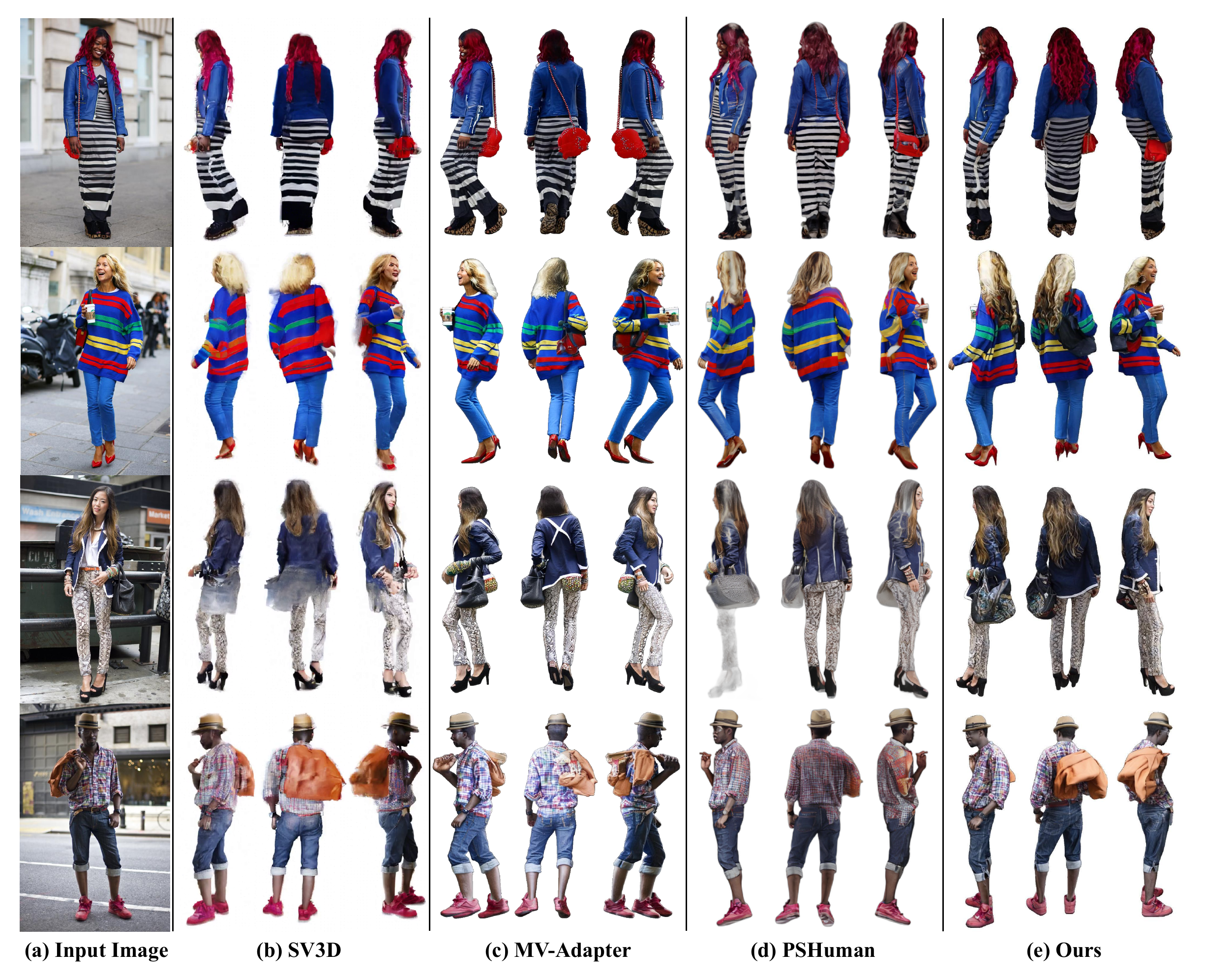}
    \vspace{-2ex}
    \caption{Visual comparison of multi-view generation results on full body images from the CCP dataset \cite{CCP}. Given the input image, we show three generated views (left, back, and right) from each method. Zoom in for more detail.} 
    \label{Fig.fullbody}
    \vspace{-2ex}
\end{figure*}

\noindent
\textbf{Numerical Comparison.}
Table \ref{tab:numerical_comparison} shows the quantitative comparison between multi-view image generation methods on in-the-wild human images. For multi-view generation of full body images, our method achieves a better result than PSHuman in all three metrics. On the CelebA dataset, our method also numerically outperforms competing methods in generating multi-view images of heads. In particular, in both full body and head images, our method achieves the highest MVReward score which is a metric that best aligns with human preferences. This suggests that our generated results are of higher quality and align better to the input image than those from competing state-of-the-art methods.

\begin{figure*}[t]
    \centering
    \includegraphics[width=.98\hsize]{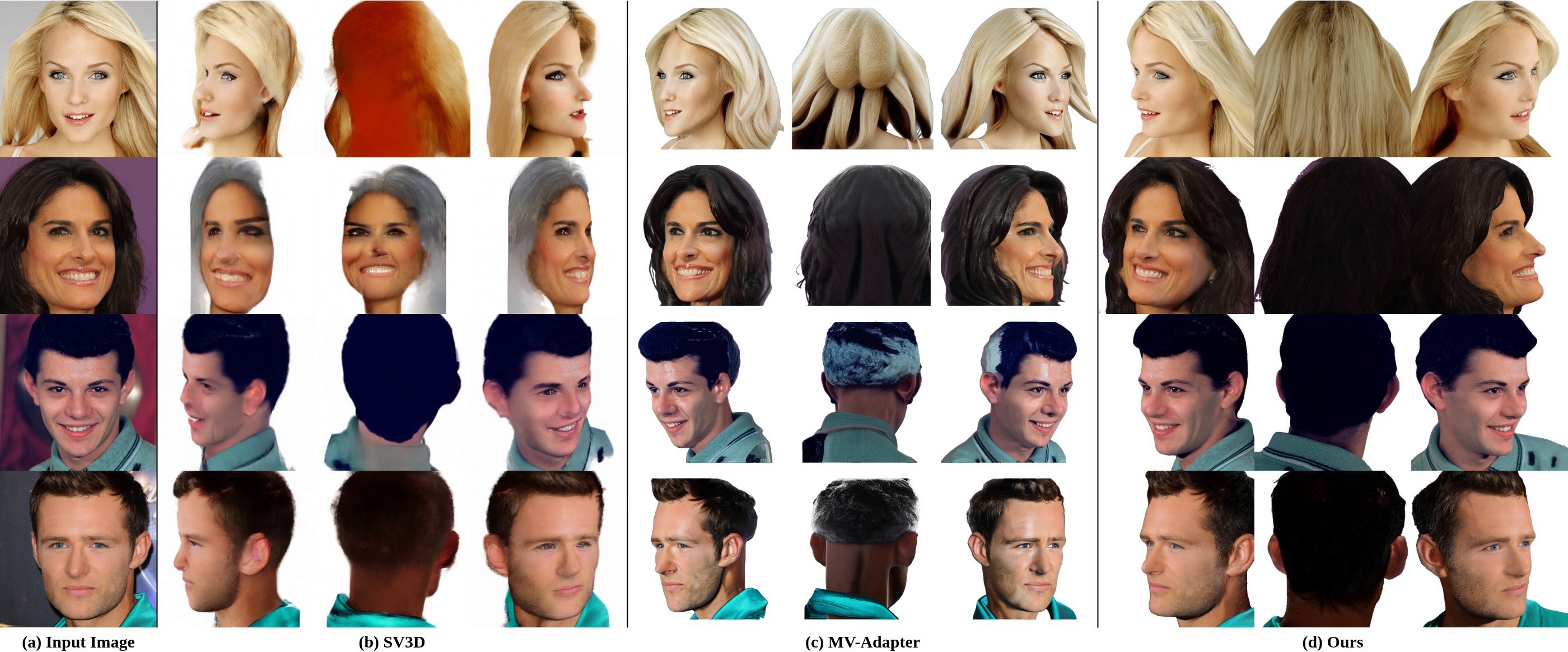}
    \vspace{-1ex}
    \caption{Visual comparison of multi-view generation results on head portrait photos from the CelebA dataset \cite{CelebAMask-HQ}. Given the input image, we show three generated views (front left, back, and front right) from each method.} 
    \label{Fig.head}
\end{figure*}

\begin{figure*}[t]
    \centering
    \includegraphics[width=.98\hsize]{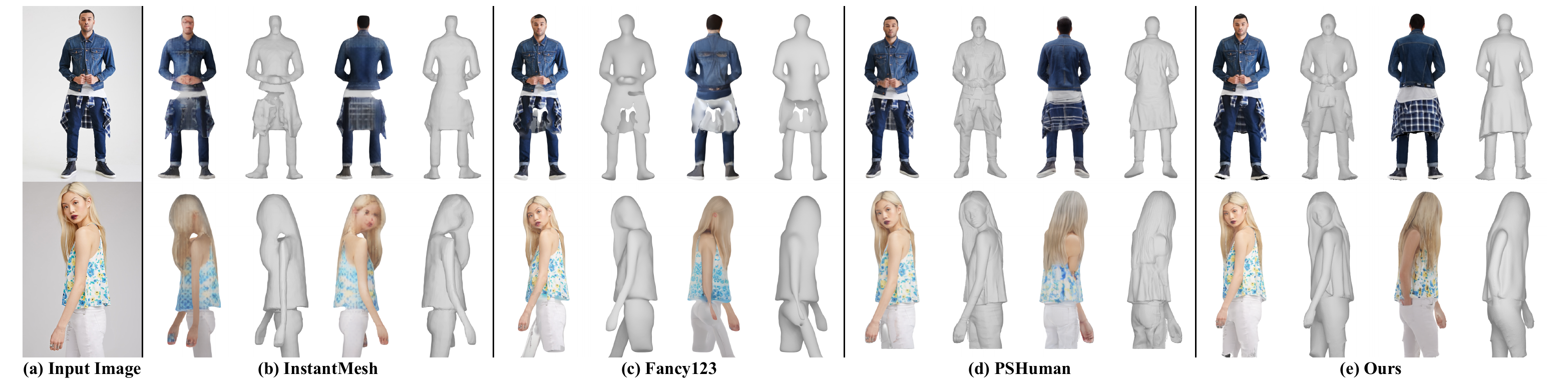}
    \vspace{-1ex}
    \caption{Visual comparison of the appearance and geometry of 3D mesh reconstruction methods on body images.} 
    \label{Fig.body_mesh}
    \vspace{-2ex}
\end{figure*}

\noindent
\textbf{Visual Comparison.}
Figure \ref{Fig.fullbody} shows the visual comparison of multi-view generation results on full body images from the CCP dataset \cite{CCP}. For each method, the figure shows three generated views: left, back, and right. For the first person, it can be seen that both SV3D and PSHuman blurred the results of the horizontal stripes on the clothing, while our results better maintains the pattern and color in each view. A similar result can be observed with the second person where the horizontal lines on the shirt is smudged in the novel views generated by the other methods. In the third image, PSHuman is unable to create a clear result of the pants from the left view, whereas our method generates realistic results in all views. In the last example, PSHuman fails to reconstruct the orange jacket held over the right shoulder while MV-Adapter seems to output more than two shoes when viewed from the side. 

From these results, it can be observed that SV3D tends to create blurry outlines and distorts the face. With PSHuman, it can generate clearer results but it lacks attention to subtle details and also creates a face that looks slightly different than the input image. In contrast, our method has great attention to detail and generates novel views that best matches the image prompt.

\begin{figure*}[t]
    \centering
    \includegraphics[width=.98\hsize]{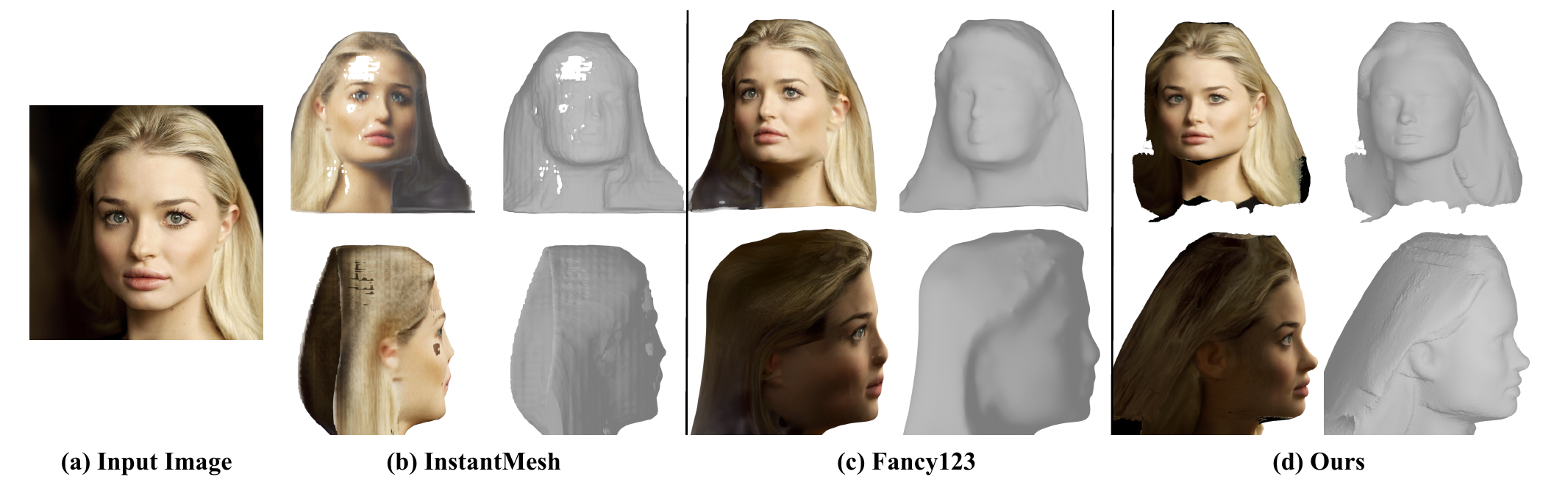}
    \vspace{-2ex}
    \caption{Visual comparison of the appearance and geometry of 3D mesh reconstruction methods on a head portrait.} 
    \label{Fig.head_mesh}
    \vspace{-2ex}
\end{figure*}

In Figure \ref{Fig.head}, we show the visual comparison of multi-view generation results on head photos from the CelebA dataset. For each input image, we show the generated results from three view angles: front left, back, and front right. With the full body images, SV3D showed slight distortions in the face area, but this is made more profound when the input image is more close-up. From the results, it can be observed that SV3D tends to create flat heads with severe artifacts where the front and back of the head appear compressed. In the second example, SV3D fails to reconstruct the back of the head and has a hair color that is inconsistent with the original image. MV-Adapter has some slight distortions, most notably in the first example where the jaw looks stretched. It also seems to create a glossy texture as seen in the second and third rows. 
In contrast, our method generates more photorealistic and consistent results that align well with the input prompt.

Examples of the 360° orbit video generated by our model as well as additional visual comparisons is provided in the supplementary.

\subsection{Single Image 3D Human Reconstruction}

\noindent
\textbf{Baselines.} 
For single image to 3D human mesh reconstruction, we conduct a visual comparison with three methods: InstantMesh \cite{instantmesh}, Fancy123 \cite{fancy123}, and PSHuman \cite{pshuman}. InstantMesh is a general mesh generation method leveraging a multi-view diffusion model \cite{zero123++} and a large reconstruction model \cite{lrm}. Fancy123 extends InstantMesh through enhancement modules with additional optimization steps for better view consistency and fidelity to the input image. PSHuman uses a cross-scale diffusion model to generate multi-view images and conducts mesh carving to reconstruct a mesh.

\noindent
\textbf{Visual Comparison.}
Figure \ref{Fig.body_mesh} shows a visual comparison of the 3D mesh reconstructed from two images from the DeepFashion \cite{deepfashion2} dataset. In the first photo, InstantMesh and Fancy123 results in hollow areas in the torso. The result from our method has smoother surface details than PSHuman, but has a better reconstruction of the left wrist and the space between the body and arms. In the second photo of a half body taken from the side, PSHuman is unable to generate the right arm in the back view. Even with a partial body, our method is able to generate a mesh with high fidelity. 

Figure \ref{Fig.head_mesh} shows a visual comparison with InstantMesh and Fancy123 on an image from the CelebA dataset. The mesh generated by InstantMesh contains holes and does not reconstruct the back of the head, while the result from Fancy123 looks stretched when viewed from the side. Compared to these methods, our method reconstructs more details such as the ear and the mouth.
 
% \subsection{Text to 3D Reconstruction}

\section{Ablation Study}
\begin{figure}[t]
    \centering
    \includegraphics[width=.98\hsize]{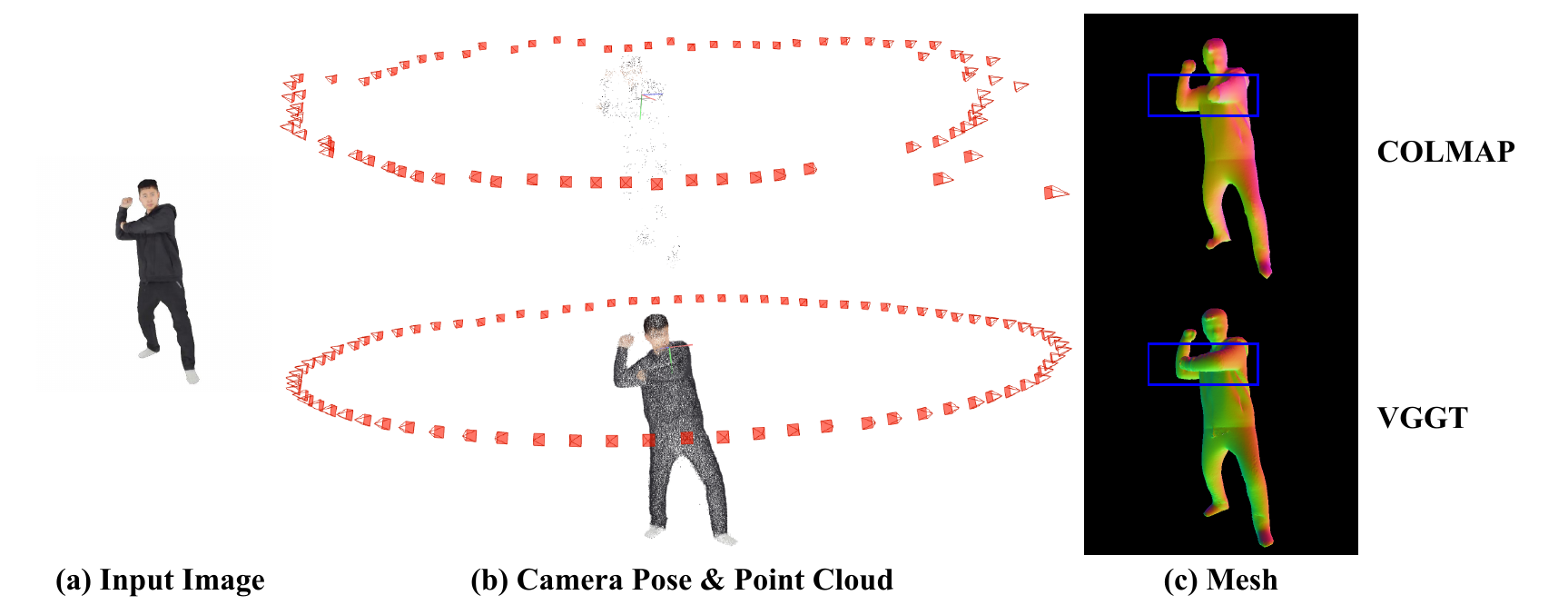}
    \caption{Comparison of 3D reconstruction when using COLMAP and VGGT for camera pose estimation.} 
    \label{Fig.colmap}
    \vspace{-2ex}
\end{figure}

\noindent
\textbf{Camera Pose Esimation.}
An alternative way to estimate the camera pose is with COLMAP \cite{colmap}. A comparison of applying COLMAP in our pipeline for camera pose estimation is shown in Figure \ref{Fig.colmap}. It can be seen that COLMAP outputs a much sparser point cloud with more discontinuous camera trajectories. This also results in a final mesh where the left arm is missing. With VGGT, we obtain a denser point cloud with accurate camera poses, resulting in a better 3D reconstruction.

% \subsection{Mesh Initialization}
% A simple and general approach for initializing the mesh for optimization is using a sphere as opposed to applying Poisson Surface Reconstruction on the predicted point cloud. While this method generally works, Figure \ref{Fig.mesh_init} shows a failure case when it is unable to create the space between the arms. Furthermore, as the Poisson mesh already has a rough structure of the body, it requires fewer optimization steps than when initialized with a sphere. 

% \begin{figure}[t]
%     \centering
%     \includegraphics[width=.98\hsize]{figures/mesh_init.pdf}
%     \caption{Comparison of mesh initialization strategy. The first and second row shows the result when initializing with a sphere and Poisson Surface Reconstruction respectively. The images show the rendered normal maps.} 
%     \label{Fig.mesh_init}
% \end{figure}

\begin{figure}[t]
    \centering
    \includegraphics[width=.98\hsize]{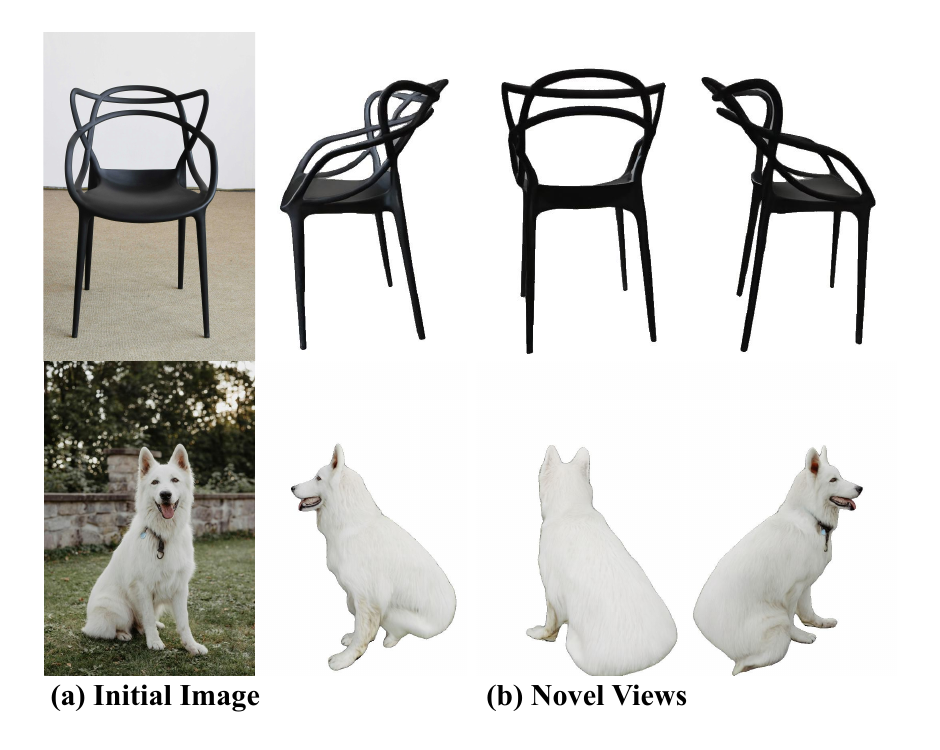}
    \vspace{-2ex}
    \caption{Novel views generated by our model on non-human images such as a chair and a dog.} 
    \label{Fig.nonhuman}
\end{figure}

\noindent
\textbf{Non-human Multi-View Generation.}
While HumanOrbit is trained on multi-view renderings of 3D human data, the base video diffusion model is trained on billions of real world videos. Recent works \cite{lora_vs_full} have shown that LoRA training forgets less than full finetuning, suggesting that our model may be capable of multi-view generation of non-human images. Figure \ref{Fig.nonhuman} displays the novel views generated bt our model for images of a chair and a dog. This result hints that the finetuned model learns accurate camera trajectories around the target object and shows the potential use cases in multi-view data generation for other objects.

\noindent
\textbf{Limitations.}
One limitation of our approach is that our model generates camera orbits at a fixed elevation. Although it captures most of the subject, it still results in some unseen areas such as the top of the head or under the chin. In future works, we can explore different camera trajectories that better capture all of the essential areas. Another limitation is the inference time. By using a large video diffusion model, it takes approximately 17 minutes to generate the full orbit video from a single input image. We can potentially improve the inference time by generating fewer frames, but initial attempts at using the same training with fewer views did not yield satisfactory results. Future works can look into other ways to reduce the computation for efficient inference.
\section{Conclusion}
In this paper, we present a method for generating dense multi-view human images from a single input image. By finetuning a LoRA on a pre-trained video diffusion model, we validate that it can learn to accurately model 360° camera orbits around the target subject with only a few hundred 3D scans. We further propose a pipeline to leverage the generated images for 3D human mesh reconstruction, demonstrating that it can be effectively used for synthesizing realistic 3D data from a 2D image. Experimental results show that our proposed method achieves competitive performance against recent methods and shows versatility towards various types of images.
{
    \small
    \bibliographystyle{ieeenat_fullname}
    \bibliography{main}

@String(CVPR= {IEEE Conf. Comput. Vis. Pattern Recog.})

@String(TOG= {ACM Trans. Graph.})

@String(ICLR = {Int. Conf. Learn. Represent.})

@String(AAAI = {AAAI})

@String(CVPR  = {CVPR})

@String(TOG   = {ACM TOG})

@String(ICLR  = {ICLR})

@inproceedings{zero123,
  title={Zero-1-to-3: Zero-shot one image to 3d object},
  author={Liu, Ruoshi and Wu, Rundi and Van Hoorick, Basile and Tokmakov, Pavel and Zakharov, Sergey and Vondrick, Carl},
  booktitle={Proceedings of the IEEE/CVF international conference on computer vision},
  pages={9298--9309},
  year={2023}
}

@article{mvdream,
  title={Mvdream: Multi-view diffusion for 3d generation},
  author={Shi, Yichun and Wang, Peng and Ye, Jianglong and Long, Mai and Li, Kejie and Yang, Xiao},
  journal={arXiv preprint arXiv:2308.16512},
  year={2023}
}

@article{syncdreamer,
  title={Syncdreamer: Generating multiview-consistent images from a single-view image},
  author={Liu, Yuan and Lin, Cheng and Zeng, Zijiao and Long, Xiaoxiao and Liu, Lingjie and Komura, Taku and Wang, Wenping},
  journal={arXiv preprint arXiv:2309.03453},
  year={2023}
}

@inproceedings{epidiff,
  title={Epidiff: Enhancing multi-view synthesis via localized epipolar-constrained diffusion},
  author={Huang, Zehuan and Wen, Hao and Dong, Junting and Wang, Yaohui and Li, Yangguang and Chen, Xinyuan and Cao, Yan-Pei and Liang, Ding and Qiao, Yu and Dai, Bo and others},
  booktitle={Proceedings of the IEEE/CVF Conference on Computer Vision and Pattern Recognition},
  pages={9784--9794},
  year={2024}
}

@article{era3d,
  title={Era3d: High-resolution multiview diffusion using efficient row-wise attention},
  author={Li, Peng and Liu, Yuan and Long, Xiaoxiao and Zhang, Feihu and Lin, Cheng and Li, Mengfei and Qi, Xingqun and Zhang, Shanghang and Xue, Wei and Luo, Wenhan and others},
  journal={Advances in Neural Information Processing Systems},
  volume={37},
  pages={55975--56000},
  year={2024}
}

@inproceedings{meat,
  title={Meat: Multiview diffusion model for human generation on megapixels with mesh attention},
  author={Wang, Yuhan and Hong, Fangzhou and Yang, Shuai and Jiang, Liming and Wu, Wayne and Loy, Chen Change},
  booktitle={Proceedings of the Computer Vision and Pattern Recognition Conference},
  pages={11297--11306},
  year={2025}
}

@inproceedings{pippo,
  title={Pippo: High-resolution multi-view humans from a single image},
  author={Kant, Yash and Weber, Ethan and Kim, Jin Kyu and Khirodkar, Rawal and Zhaoen, Su and Martinez, Julieta and Gilitschenski, Igor and Saito, Shunsuke and Bagautdinov, Timur},
  booktitle={Proceedings of the Computer Vision and Pattern Recognition Conference},
  pages={16418--16429},
  year={2025}
}

@article{human4dit,
title={Human4DiT: 360-degree Human Video Generation with 4D Diffusion Transformer},
author={Shao, Ruizhi and Pang, Youxin and Zheng, Zerong and Sun, Jingxiang and Liu, Yebin},
journal={ACM Transactions on Graphics (TOG)},
volume={43},
number={6},
articleno={},
year={2024}, publisher={ACM New York, NY, USA}
}

@inproceedings{pifu,
  title={Pifu: Pixel-aligned implicit function for high-resolution clothed human digitization},
  author={Saito, Shunsuke and Huang, Zeng and Natsume, Ryota and Morishima, Shigeo and Kanazawa, Angjoo and Li, Hao},
  booktitle={Proceedings of the IEEE/CVF international conference on computer vision},
  pages={2304--2314},
  year={2019}
}

@inproceedings{pifuhd,
  title={Pifuhd: Multi-level pixel-aligned implicit function for high-resolution 3d human digitization},
  author={Saito, Shunsuke and Simon, Tomas and Saragih, Jason and Joo, Hanbyul},
  booktitle={Proceedings of the IEEE/CVF conference on computer vision and pattern recognition},
  pages={84--93},
  year={2020}
}

@inproceedings{icon,
  title={Icon: Implicit clothed humans obtained from normals},
  author={Xiu, Yuliang and Yang, Jinlong and Tzionas, Dimitrios and Black, Michael J},
  booktitle={2022 IEEE/CVF Conference on Computer Vision and Pattern Recognition (CVPR)},
  pages={13286--13296},
  year={2022},
  organization={IEEE}
}

@article{pamir,
  title={Pamir: Parametric model-conditioned implicit representation for image-based human reconstruction},
  author={Zheng, Zerong and Yu, Tao and Liu, Yebin and Dai, Qionghai},
  journal={IEEE transactions on pattern analysis and machine intelligence},
  volume={44},
  number={6},
  pages={3170--3184},
  year={2021},
  publisher={IEEE}
}

@article{hunyuanvideo,
  title={Hunyuanvideo: A systematic framework for large video generative models},
  author={Kong, Weijie and Tian, Qi and Zhang, Zijian and Min, Rox and Dai, Zuozhuo and Zhou, Jin and Xiong, Jiangfeng and Li, Xin and Wu, Bo and Zhang, Jianwei and others},
  journal={arXiv preprint arXiv:2412.03603},
  year={2024}
}

@article{cogvideox,
  title={Cogvideox: Text-to-video diffusion models with an expert transformer},
  author={Yang, Zhuoyi and Teng, Jiayan and Zheng, Wendi and Ding, Ming and Huang, Shiyu and Xu, Jiazheng and Yang, Yuanming and Hong, Wenyi and Zhang, Xiaohan and Feng, Guanyu and others},
  journal={arXiv preprint arXiv:2408.06072},
  year={2024}
}

@article{svd,
  title={Stable video diffusion: Scaling latent video diffusion models to large datasets},
  author={Blattmann, Andreas and Dockhorn, Tim and Kulal, Sumith and Mendelevitch, Daniel and Kilian, Maciej and Lorenz, Dominik and Levi, Yam and English, Zion and Voleti, Vikram and Letts, Adam and others},
  journal={arXiv preprint arXiv:2311.15127},
  year={2023}
}

@inproceedings{pshuman,
  title={Pshuman: Photorealistic single-image 3d human reconstruction using cross-scale multiview diffusion and explicit remeshing},
  author={Li, Peng and Zheng, Wangguandong and Liu, Yuan and Yu, Tao and Li, Yangguang and Qi, Xingqun and Chi, Xiaowei and Xia, Siyu and Cao, Yan-Pei and Xue, Wei and others},
  booktitle={Proceedings of the Computer Vision and Pattern Recognition Conference},
  pages={16008--16018},
  year={2025}
}

@inproceedings{sv3d,
  title={Sv3d: Novel multi-view synthesis and 3d generation from a single image using latent video diffusion},
  author={Voleti, Vikram and Yao, Chun-Han and Boss, Mark and Letts, Adam and Pankratz, David and Tochilkin, Dmitry and Laforte, Christian and Rombach, Robin and Jampani, Varun},
  booktitle={European Conference on Computer Vision},
  pages={439--457},
  year={2024},
  organization={Springer}
}

@inproceedings{mv-adapter,
  title={Mv-adapter: Multi-view consistent image generation made easy},
  author={Huang, Zehuan and Guo, Yuan-Chen and Wang, Haoran and Yi, Ran and Ma, Lizhuang and Cao, Yan-Pei and Sheng, Lu},
  booktitle={Proceedings of the IEEE/CVF International Conference on Computer Vision},
  pages={16377--16387},
  year={2025}
}

@article{sdxl,
  title={Sdxl: Improving latent diffusion models for high-resolution image synthesis},
  author={Podell, Dustin and English, Zion and Lacey, Kyle and Blattmann, Andreas and Dockhorn, Tim and M{\"u}ller, Jonas and Penna, Joe and Rombach, Robin},
  journal={arXiv preprint arXiv:2307.01952},
  year={2023}
}

@inproceedings{spinmeround,
  title={SpinMeRound: Consistent Multi-View Identity Generation Using Diffusion Models},
  author={Galanakis, Stathis and Lattas, Alexandros and Moschoglou, Stylianos and Kainz, Bernhard and Zafeiriou, Stefanos},
  booktitle={Proceedings of the IEEE/CVF International Conference on Computer Vision},
  pages={14346--14356},
  year={2025}
}

@inproceedings{fancy123,
  title={Fancy123: One Image to High-Quality 3D Mesh Generation via Plug-and-Play Deformation},
  author={Yu, Qiao and Li, Xianzhi and Tang, Yuan and Han, Xu and Hu, Long and Hao, Yixue and Chen, Min},
  booktitle={Proceedings of the Computer Vision and Pattern Recognition Conference},
  pages={595--604},
  year={2025}
}

@article{instantmesh,
  title={InstantMesh: Efficient 3D Mesh Generation from a Single Image with Sparse-view Large Reconstruction Models},
  author={Xu, Jiale and Cheng, Weihao and Gao, Yiming and Wang, Xintao and Gao, Shenghua and Shan, Ying},
  journal={arXiv preprint arXiv:2404.07191},
  year={2024}
}

@article{zero123++,
  title={Zero123++: a single image to consistent multi-view diffusion base model},
  author={Shi, Ruoxi and Chen, Hansheng and Zhang, Zhuoyang and Liu, Minghua and Xu, Chao and Wei, Xinyue and Chen, Linghao and Zeng, Chong and Su, Hao},
  journal={arXiv preprint arXiv:2310.15110},
  year={2023}
}

@article{lrm,
  title={Lrm: Large reconstruction model for single image to 3d},
  author={Hong, Yicong and Zhang, Kai and Gu, Jiuxiang and Bi, Sai and Zhou, Yang and Liu, Difan and Liu, Feng and Sunkavalli, Kalyan and Bui, Trung and Tan, Hao},
  journal={arXiv preprint arXiv:2311.04400},
  year={2023}
}

@inproceedings{CCP,
  title={Clothing co-parsing by joint image segmentation and labeling},
  author={Yang, Wei and Luo, Ping and Lin, Liang},
  booktitle={Proceedings of the IEEE conference on computer vision and pattern recognition},
  pages={3182--3189},
  year={2014}
}

@inproceedings{CelebAMask-HQ,
  title={Maskgan: Towards diverse and interactive facial image manipulation},
  author={Lee, Cheng-Han and Liu, Ziwei and Wu, Lingyun and Luo, Ping},
  booktitle={Proceedings of the IEEE/CVF conference on computer vision and pattern recognition},
  pages={5549--5558},
  year={2020}
}

@inproceedings{deepfashion1,
 author = {Liu, Ziwei and Luo, Ping and Qiu, Shi and Wang, Xiaogang and Tang, Xiaoou},
 title = {DeepFashion: Powering Robust Clothes Recognition and Retrieval with Rich Annotations},
 booktitle = {Proceedings of IEEE Conference on Computer Vision and Pattern Recognition (CVPR)},
 month = {June},
 year = {2016}
 }

@article{deepfashion2,
  title={Text2Human: Text-Driven Controllable Human Image Generation},
  author={Jiang, Yuming and Yang, Shuai and Qiu, Haonan and Wu, Wayne and Loy, Chen Change and Liu, Ziwei},
  journal={ACM Transactions on Graphics (TOG)},
  volume={41},
  number={4},
  articleno={162},
  pages={1--11},
  year={2022},
  publisher={ACM New York, NY, USA},
  doi={10.1145/3528223.3530104},
}

@inproceedings{clip,
  title={Learning transferable visual models from natural language supervision},
  author={Radford, Alec and Kim, Jong Wook and Hallacy, Chris and Ramesh, Aditya and Goh, Gabriel and Agarwal, Sandhini and Sastry, Girish and Askell, Amanda and Mishkin, Pamela and Clark, Jack and others},
  booktitle={International conference on machine learning},
  pages={8748--8763},
  year={2021},
  organization={PmLR}
}

@inproceedings{met3r,
  title={Met3r: Measuring multi-view consistency in generated images},
  author={Asim, Mohammad and Wewer, Christopher and Wimmer, Thomas and Schiele, Bernt and Lenssen, Jan Eric},
  booktitle={Proceedings of the Computer Vision and Pattern Recognition Conference},
  pages={6034--6044},
  year={2025}
}

@inproceedings{mvreward,
  title={Mvreward: Better aligning and evaluating multi-view diffusion models with human preferences},
  author={Wang, Weitao and Xu, Haoran and Yang, Yuxiao and Liu, Zhifang and Meng, Jun and Wang, Haoqian},
  booktitle={Proceedings of the AAAI Conference on Artificial Intelligence},
  volume={39},
  number={8},
  pages={7898--7906},
  year={2025}
}

@inproceedings{dust3r,
  title={Dust3r: Geometric 3d vision made easy},
  author={Wang, Shuzhe and Leroy, Vincent and Cabon, Yohann and Chidlovskii, Boris and Revaud, Jerome},
  booktitle={Proceedings of the IEEE/CVF Conference on Computer Vision and Pattern Recognition},
  pages={20697--20709},
  year={2024}
}

@inproceedings{dino,
  title={Emerging properties in self-supervised vision transformers},
  author={Caron, Mathilde and Touvron, Hugo and Misra, Ishan and J{\'e}gou, Herv{\'e} and Mairal, Julien and Bojanowski, Piotr and Joulin, Armand},
  booktitle={Proceedings of the IEEE/CVF international conference on computer vision},
  pages={9650--9660},
  year={2021}
}

@article{dinov2,
  title={Dinov2: Learning robust visual features without supervision},
  author={Oquab, Maxime and Darcet, Timoth{\'e}e and Moutakanni, Th{\'e}o and Vo, Huy and Szafraniec, Marc and Khalidov, Vasil and Fernandez, Pierre and Haziza, Daniel and Massa, Francisco and El-Nouby, Alaaeldin and others},
  journal={arXiv preprint arXiv:2304.07193},
  year={2023}
}

@article{featup,
  title={Featup: A model-agnostic framework for features at any resolution},
  author={Fu, Stephanie and Hamilton, Mark and Brandt, Laura and Feldman, Axel and Zhang, Zhoutong and Freeman, William T},
  journal={arXiv preprint arXiv:2403.10516},
  year={2024}
}

@article{lora_vs_full,
  title={Lora vs full fine-tuning: An illusion of equivalence},
  author={Shuttleworth, Reece and Andreas, Jacob and Torralba, Antonio and Sharma, Pratyusha},
  journal={arXiv preprint arXiv:2410.21228},
  year={2024}
}

@article{wan,
  title={Wan: Open and advanced large-scale video generative models},
  author={Wan, Team and Wang, Ang and Ai, Baole and Wen, Bin and Mao, Chaojie and Xie, Chen-Wei and Chen, Di and Yu, Feiwu and Zhao, Haiming and Yang, Jianxiao and others},
  journal={arXiv preprint arXiv:2503.20314},
  year={2025}
}

@article{umt5,
  title={Unimax: Fairer and more effective language sampling for large-scale multilingual pretraining},
  author={Chung, Hyung Won and Constant, Noah and Garcia, Xavier and Roberts, Adam and Tay, Yi and Narang, Sharan and Firat, Orhan},
  journal={arXiv preprint arXiv:2304.09151},
  year={2023}
}

@inproceedings{dit,
  title={Scalable diffusion models with transformers},
  author={Peebles, William and Xie, Saining},
  booktitle={Proceedings of the IEEE/CVF international conference on computer vision},
  pages={4195--4205},
  year={2023}
}

@article{lora,
  title={Lora: Low-rank adaptation of large language models.},
  author={Hu, Edward J and Shen, Yelong and Wallis, Phillip and Allen-Zhu, Zeyuan and Li, Yuanzhi and Wang, Shean and Wang, Lu and Chen, Weizhu and others},
  journal={ICLR},
  volume={1},
  number={2},
  pages={3},
  year={2022}
}

@misc{RenderPeople,
    author= {{Renderpeople}},
    year  = {2023},
    note  = {\url{https://renderpeople.com/3d-people}}
}

@article{normalcrafter,
  title={Normalcrafter: Learning temporally consistent normals from video diffusion priors},
  author={Bin, Yanrui and Hu, Wenbo and Wang, Haoyuan and Chen, Xinya and Wang, Bing},
  journal={arXiv preprint arXiv:2504.11427},
  year={2025}
}

@inproceedings{vggt,
  title={Vggt: Visual geometry grounded transformer},
  author={Wang, Jianyuan and Chen, Minghao and Karaev, Nikita and Vedaldi, Andrea and Rupprecht, Christian and Novotny, David},
  booktitle={Proceedings of the Computer Vision and Pattern Recognition Conference},
  pages={5294--5306},
  year={2025}
}

@article{continuous,
  title={Continuous remeshing for inverse rendering},
  author={Palfinger, Werner},
  journal={Computer Animation and Virtual Worlds},
  volume={33},
  number={5},
  pages={e2101},
  year={2022},
  publisher={Wiley Online Library}
}

@article{unique3d,
  title={Unique3d: High-quality and efficient 3d mesh generation from a single image},
  author={Wu, Kailu and Liu, Fangfu and Cai, Zhihan and Yan, Runjie and Wang, Hanyang and Hu, Yating and Duan, Yueqi and Ma, Kaisheng},
  journal={Advances in Neural Information Processing Systems},
  volume={37},
  pages={125116--125141},
  year={2024}
}

@inproceedings{craftsman3d,
  title={CraftsMan3D: High-fidelity Mesh Generation with 3D Native Diffusion and Interactive Geometry Refiner},
  author={Li, Weiyu and Liu, Jiarui and Yan, Hongyu and Chen, Rui and Liang, Yixun and Chen, Xuelin and Tan, Ping and Long, Xiaoxiao},
  booktitle={Proceedings of the Computer Vision and Pattern Recognition Conference},
  pages={5307--5317},
  year={2025}
}

@article{infinihuman,
  author    = {Xue, Yuxuan and Xie, Xianghui and Kostyrko, Margaret and Pons-Moll, Gerard},
  title     = {InfiniHuman: Infinite 3D Human Creation with Precise Control},
  booktitle = {SIGGRAPH Asia 2025 Conference Papers},
  year      = {2025},
}

@inproceedings{poisson,
  title={Poisson surface reconstruction},
  author={Kazhdan, Michael and Bolitho, Matthew and Hoppe, Hugues},
  booktitle={Proceedings of the fourth Eurographics symposium on Geometry processing},
  volume={7},
  number={4},
  year={2006}
}

@incollection{smpl,
  title={SMPL: A skinned multi-person linear model},
  author={Loper, Matthew and Mahmood, Naureen and Romero, Javier and Pons-Moll, Gerard and Black, Michael J},
  booktitle={Seminal Graphics Papers: Pushing the Boundaries, Volume 2},
  pages={851--866},
  year={2023}
}

@inproceedings{smplx,
    title = {Expressive Body Capture: 3D Hands, Face, and Body from a Single Image},
    author = {Pavlakos, Georgios and Choutas, Vasileios and Ghorbani, Nima and Bolkart, Timo and Osman, Ahmed A. A. and Tzionas, Dimitrios and Black, Michael J.},
    booktitle = {Proceedings IEEE Conf. on Computer Vision and Pattern Recognition (CVPR)},
    year = {2019}
}

@inproceedings{colmap,
    author={Sch\"{o}nberger, Johannes Lutz and Frahm, Jan-Michael},
    title={Structure-from-Motion Revisited},
    booktitle={Conference on Computer Vision and Pattern Recognition (CVPR)},
    year={2016},
}

@misc{stablediffusion,
      title={High-Resolution Image Synthesis with Latent Diffusion Models}, 
      author={Robin Rombach and Andreas Blattmann and Dominik Lorenz and Patrick Esser and Björn Ommer},
      year={2021},
      eprint={2112.10752},
      archivePrefix={arXiv},
      primaryClass={cs.CV}
}

@inproceedings{objaverse,
  title={Objaverse: A universe of annotated 3d objects},
  author={Deitke, Matt and Schwenk, Dustin and Salvador, Jordi and Weihs, Luca and Michel, Oscar and VanderBilt, Eli and Schmidt, Ludwig and Ehsani, Kiana and Kembhavi, Aniruddha and Farhadi, Ali},
  booktitle={Proceedings of the IEEE/CVF conference on computer vision and pattern recognition},
  pages={13142--13153},
  year={2023}
}

@inproceedings{mvimgnet,
  title={Mvimgnet: A large-scale dataset of multi-view images},
  author={Yu, Xianggang and Xu, Mutian and Zhang, Yidan and Liu, Haolin and Ye, Chongjie and Wu, Yushuang and Yan, Zizheng and Zhu, Chenming and Xiong, Zhangyang and Liang, Tianyou and others},
  booktitle={Proceedings of the IEEE/CVF conference on computer vision and pattern recognition},
  pages={9150--9161},
  year={2023}
}

@article{weng2024template,
  title={Template-free single-view 3d human digitalization with diffusion-guided lrm},
  author={Weng, Zhenzhen and Liu, Jingyuan and Tan, Hao and Xu, Zhan and Zhou, Yang and Yeung-Levy, Serena and Yang, Jimei},
  journal={arXiv preprint arXiv:2401.12175},
  year={2024}
}
}


\end{document}